\documentclass{easychair}
\usepackage{amsmath}
\usepackage{latexsym}
\usepackage{balance}
\usepackage{graphicx}
\usepackage{url}
\usepackage{amsmath}
\usepackage{float}
\usepackage{graphicx}
\usepackage{graphics}
\usepackage{pdfpages}
\usepackage{vmargin}
\usepackage{subfigure}
\usepackage{hyperref}
\usepackage{ifthen}
\usepackage{theorem}
\usepackage{listings} 
\usepackage{amssymb} 
\usepackage[latin1]{inputenc}

\title{Gelisp: A Library to
  Represent Musical CSPs and Search Strategies}

\author{
Mauricio Toro\inst{1}
\and
Camilo Rueda\inst{2}
\and
Carlos Agón\inst{3}
\and \\
Gérard Assayag\inst{3}
}

\institute{
  Universidad Eafit,
  Medellin, Colombia\\
\and
   Pontificia Universidad Javeriana Cali,
   Cali, Colombia\\
\and
   IRCAM,
   Paris, France\\
 }

\begin{document}
\bibliographystyle{abbrv}
\maketitle


\begin{abstract}
In this paper we present Gelisp, a new library to represent musical Constraint Satisfaction Problems and
search strategies intuitively. Gelisp has two interfaces,
a command-line one for Common Lisp and a graphical one for OpenMusic. Using Gelisp,  
we solved a problem of automatic music generation proposed by composer Michael Jarrell and we found solutions for 
the All-interval series.
\end{abstract}

\section{Introduction} 
A \textit{Constraint Satisfaction Problem} (CSP) is a formalism to represent combinatorial problems. To solve a CSP we need to find objects that satisfy a number of constraints (i.e., criteria
over those variables).
CSPs provide a declarative way to represent combinatorial problems, specifying  
cons-traints instead of a sequence of steps to   
find the solution (as used in imperative programming).
Additionally, it is possible to specify 
 strategies to choo-se between branches during search.  

CSPs in computer music can be used to solve harmonic, rhythmic or melodic
problems. In addition, they can be used for automatic generation of
musical structures satisfying a set of rules.
For instance, we can find solutions for the \textit{All-interval series} \cite{allinterval}, where
we need to find a sequence of 12 different pitches with 12 different intervals.

In order to solve a CSP, we can use constraint programming languages such 
as Prolog or Mozart-Oz \cite{VanRoyHaridi:2004}. In order to solve a CSP, those languages use a \textit{Constraint Solving Library (CSL)} such as Gecode \cite{fastprop}. CLSs are usually written
in C++.


\subsection{The problem}

Using traditional CSL's or programming languages to solve CSPs is
time-demanding and it is intended for specialized users because they usually
require deep knowledge on C++ or logic programming. 
This makes these tools
often unpractical to specify musical CSPs. 
Furthermore, these tools do not provide a representation
for musical data structures.


\subsection{Our solution} 
\textit{Gelisp}\footnote{http://gelisp.sourceforge.net/} is a wrapper
for Gecode to Common Lisp. 
\textit{Gelisp} was originally developed by Rueda in 2006 and we 
modified it to work with current version of Gecode.
Furthermore, we added support to model CSPs and search strategies graphically on \textit{OpenMusic (OM)} \cite{OM98}.
In addition, \textit{Gelisp} can take advantage of the musical data structures and functions 
defined for OM.

The novelty of \textit{Gelisp} is to provide a graphical
representation for search strategies (e.g., Depth\\ First Search) and global 
constraints (e.g., ``all the intervals of a sequence
must be different''), based on
an efficient CSL. 



\subsection{Related Work}
Several graphical CSLs for OM have been
developed in the last decade. 
Situation \cite{situation} generates music based on constraints, OmRc \cite{omrc}  finds structures corresponding to rhythmical constraints, OmClouds \cite{omclouds} finds 
 approximated solutions to a CSP, and OMBacktrack (\url{http://www.ircam.fr/}\\ \url{equipes/repmus/}) is a wrapper for the CSL Screamer \cite{screamer} (a CSL written on Lisp).





A  graphical CSL
to solve musical CSPs should be able to 
setup search strategies in a graphical way,
 post multiple kinds of constraints graphically without declaring explicitly loops and recursion, 
and solve the problem using state-of-art algorithms.


Unfortunately, OmRC and OmSituation are designed to
solve specific problems. OmBacktrack is no longer available
for current versions of OM. Finally, OmClouds does not guarantee a solution satisfying all the constraints (i.e., a complete solution).

\section{Gecode}
Gecode is a Constraint Solving Library (CSL) written in C++. 
Gecode provides a \textit{propagator} for each type
of constraint. Propagators translate 
a constraint into basic constraints supplying the same information.
Basic (finite domain) constraints have the form $x \in [a..b]$.
For instance, in a store (i.e., a set with all the constraints asserted) 
containing {$pitch_1 \in [36..72]$ and $pitch_2 \in [60..80]$}, a
propagator for the constraint $pitch_1 > pitch_2 + 2$ would add 
constraints $pitch_1 \in [63..72]$ and $pitch_2 \in [60..69]$.

As described in the above example, the action of propagators
ends up narrowing down the set of possible values for each
variable. This, however, does not guarantee that it will eventually
be inferred a single value for each variable. Gecode 
thus include \textit{search engines}. The purpose of a search
engine is to choose additional basic constraints to add into the
store until all variables have reduced their domain to 
a single value. Using them we can find one, many, or all the solutions
for a CSP.

Gecode works
on different operating systems and it will be used as the CSL
for Mozart-Oz, therefore it is very likely to be maintained
for a long time. Furthermore, it provides an extensible API, allowing
the user to create new propagators and user-defined search engines. For instance, we can extend Gecode
to reason about trees and graphs, which are useful in musical CSPs. 

\section{Gelisp}
\textit{Gelisp} provides an interface for Common Lisp and another 
for OM. In \textit{Gelisp}, sequences of variables are represented by lists, as opposed 
to Gecode, where they are represented by arrays. This makes the power
of list processing (provided by Lisp and OM) available for \textit{Gelisp} users.

\subsection{Interface for Common Lisp}

To solve a problem using this interface,
we need to write a script. A script is a function
to 
define the problem variables and their domains (the possible values that a variable can take), post constraints
over the variables, and setup a search strategy.

This interface allows the user to call
most of Gecode propagators for both, Finite Domain (FD)
and Finite Set (FS) constraints. Basic  FD constraints deal
with expressions of the form $x \in R$, where $R$ is a range or a 
set of ranges of integers. On the other hand, FS constraints deal
with expressions among sets of FD variables. In what follows, we present some
propagators that \textit{Gelisp} provides for FD and FS.


\textit{Gelisp} provides FD propagators for defining domains (e.g., $Domain(X) = [2,5]$), equalities and
inequalities (e.g., $ X + Y < Z$), cardinality (e.g., 1 occurs two times in $[X Y Z]$),  boolean constraints, regular
expression constraints and the all-distinct constraint. The all-distinct constraint makes the elements of a sequence pairwise different. 

On the other hand, for FS we provide constraints for defining domains (e.g., $V  \subseteq{ \{1, 2, 3 \} }$) and set relations (e.g., $X \subset{ A \cup B  }$).

In addition, \textit{Gelisp} includes two search engines,  
Depth Search First (DSF) and Branch-and-bound (BAB). 
The DFS engine works by choosing some variable, then a value for that variable, if this 
does not succeed (a constraint does not hold) then choo-ses another value. If the value 
succeed, then choo-ses another variable, then a value for it, etc. 

The BAB engine works 
in a similar way, but solutions are computed in such a way that each subsequent solution 
increases or decreases the value of some user specified FD variable. 
Both engines can be used
for both FS and FD. In addition, we can define search heuristics for value (i.e., the order to assign a value to a variable) and variable order (i.e., the order to choose a variable). These heuristics are parameters for the search engines.



\subsection{Graphical Interface for OpenMusic}
Instead of writing a script, in the graphical 
interface we represent a program with a special
patch, called \textit{CSP patch}. A patch is a visual algorithm, in which boxes represent functional
calls, and connections are functional compositions. 
Inside a \textit{CSP patch}, we can place special boxes to
define a constraint in
the CSP, variable and value heuristics, the variable to be optimized during the search, and 
a time limit in the search.

For instance, we provide a variety of boxes to 
represent simple 
constraints (e.g., $a = 2$) and 
global constraints (e.g., ``all the intervals
from a sequence must be different'').

Using the graphical interface we can express a variety of problems declaratively with global constraints.
Global constraints have parameters.
For instance, the graphical box to find the intervals of
a list ``$x \rightarrow dx$'' has a parameter to choose among absolute, non-absolute,
or modulo $n$ intervals (calculated as $(V_{i+1} - V_i) \% n$). Additionally, it has a parameter to post an \textit{all-distinct} constraint over the intervals.

Moreover, the output of a CSP
patch can be connected to a box to find one solution or a box  
to find $n$ the solutions

\section{Applications}

In this section, we describe both, an intuitive and formal definition of two
CSPs and we explain how to solve them with \textit{Gelisp}.
Formally, a CSP is triple $<X,D,C>$, where $X$ is a set of variables, $D$ is the domain for
each variable, and C is a set of constraints (read as conjunction) over the variables.

\subsection{All-interval series}

\begin{figure}[h!]
  \begin{center}
    \mbox{
     
{\includegraphics[width=\columnwidth]{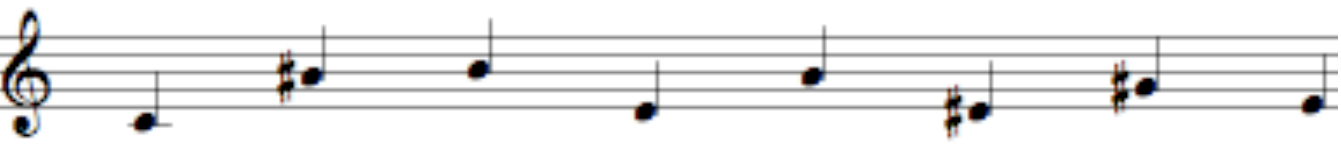}}  
 
      }
    \caption{An all-interval serie for $n=12$}
    \label{fig:allintervals-ex}
  \end{center}
\end{figure}

In this problem, 
we need to find a sequence of 12 different pitches with 12 different intervals (fig. \ref{fig:allintervals-ex}). 
This problem can be generalized to find $n$
different pitches with $n$ different intervals equivalent under inversion \footnote{For instance, an interval C-E is equivalent to E-C.}. For instance, a value of $n=24$ represents the
\textit{all-interval series} for microtones.

Therefore, a solution to this CSP is a sequence of $n$ pairwise different variables
with domain $[1..n]$, where all modulo $n$ intervals of the sequence are pairwise different. We give bellow a formalization of this problem \\ \\
\textbf{Variables}: $V_1$ ... $V_n$ \\
\textbf{Domains}: $[1..n]$ ...  $[1..n]$ \\
\textbf{Constraints}:
\begin{itemize}
\item  $C_1$ alldiff$(V)$
\item  $C_2$ alldiff$( (V_{i+1} - V_i) \% n, i \leq n)$
\end{itemize}

There is not a constraint over the interval
$(V_{n} - V_0$) because that interval is always six, according
to the literature. Furthermore, 
it is enough to calculate the series where $V_0 = 0$ 
because the other ones can be obtained from that one using transposition.
In addition, we know that if  $V_1..V_n$ is an all-interval  
serie, $V_n...V_1$ is also one. For those reasons, we include these 
two constraints to avoid symmetrical solutions:
\begin{itemize}
\item  $C_3$ $V_0 = 0$
\item  $C_4$ $V_0 < V_n$ 
\end{itemize}

\begin{figure}[h!]
  \begin{center}
    \mbox{
     
{\includegraphics[width=\columnwidth]{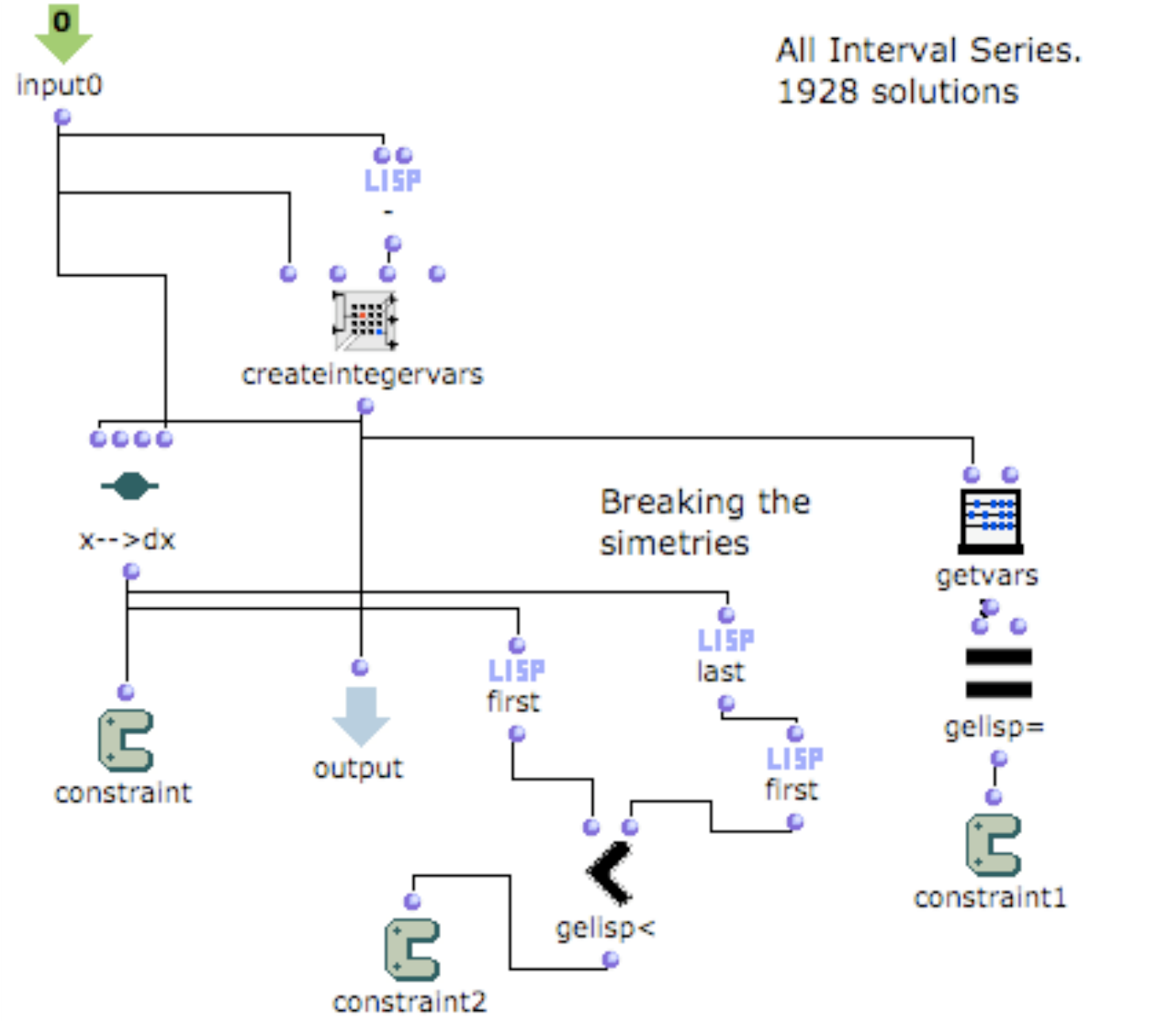}}  
 
      }
    \caption{All-interval Series CSP on OM}
    \label{fig:allintervals}
  \end{center}
\end{figure}

We represent graphically this CSP (fig. \ref{fig:allintervals}) with a box to create $n$ all-different variables with domain $[1..n]$, an $x \rightarrow dx$ box for $C_2$ with an all-different parameter
, an \textit{equality} box for $C_3$, and  an \textit{inequality} box for $C_4$.

\subsection{Jarrell CSP}
Composer Michael Jarrell proposed an idea for automatic music generation \cite{jarrell}. The goal is to generate
a sequence of $n$ notes.
There is a fix number of occurrences $OM_{1}$...$OM_{A}$ for each sequences of intervals (called motives) $M_{1}$...$M_{A}$ over the sequence of non-absolute intervals of the output sequence. In addition, each note of the output sequence belongs to a Chord $Ch$.
Moreover, 
the first $L_1$ and the
last note $L_2$ of the output sequence are fixed. We give bellow a formalization of this problem \\ \\
\textbf{Inputs}: 
\begin{itemize}
\item Motives [$M_{1}$...$M_{A}$], Limits $L_{1}$ and $L_{2}$,\\Occurrences [$OM_{1}$...$OM_{A}$], Chord $Ch$
\end{itemize}
\textbf{Variables}: $V_1$ ... $V_n$\\
\textbf{Domains}: $[0..127]$ ... $[0..127]$\\
\textbf{Constraints}:
\begin{itemize}
\item$C_1$  $\forall_{1 < i < A}$ $|\{j, M_{i}$ is a subsequence of the variables' intervals that starts on $j\}| = OM_{i}$
\item$C_2$  $\forall_{1 < i < n}$ $V_i \in Ch$     
\item$C_3$ $V_1 = L_1 \wedge V_n = L_2$  
\end{itemize}

\begin{figure}[h!]
  \begin{center}
    \mbox{
     
{\includegraphics[width=\columnwidth]{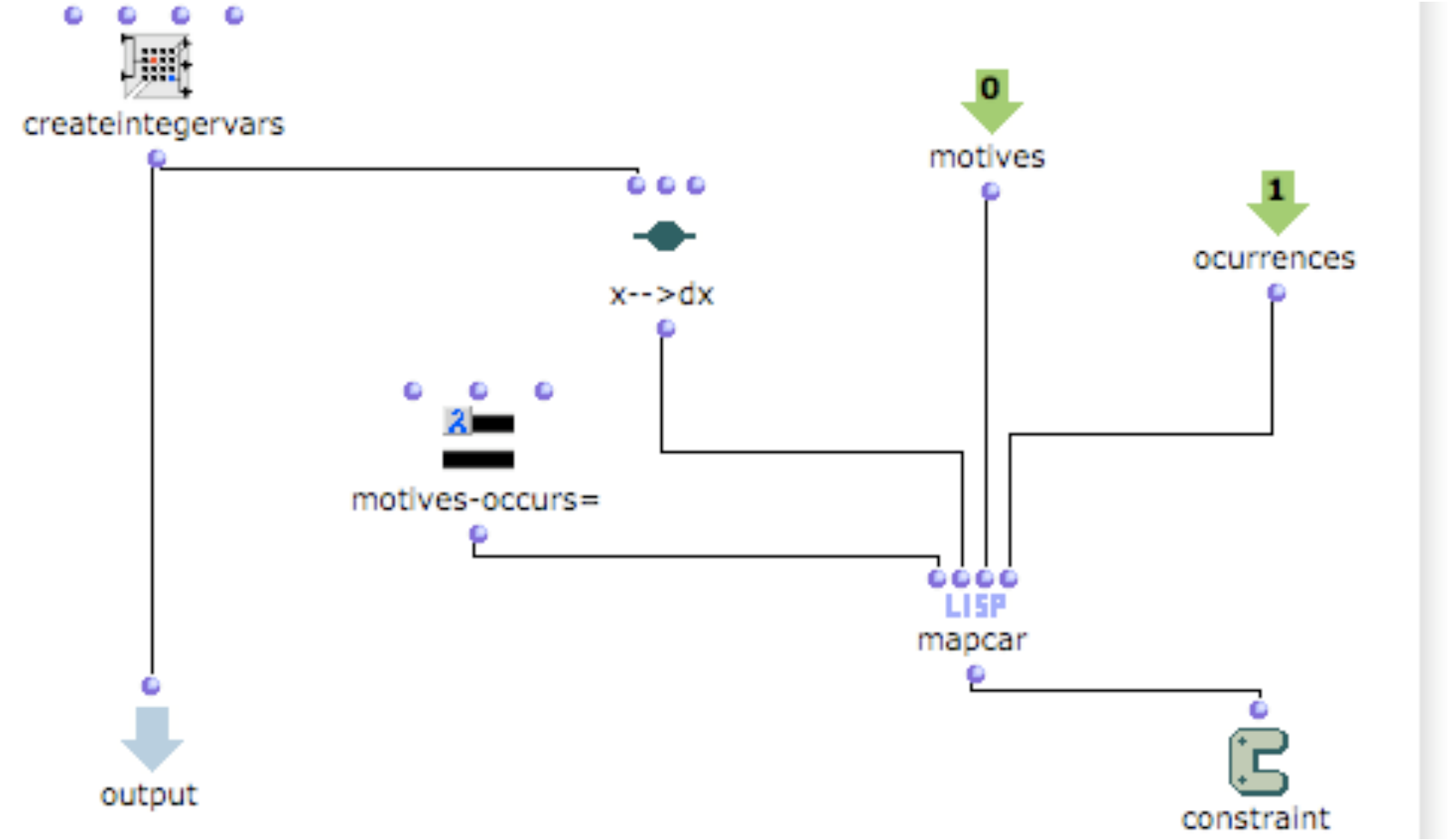}}  
 
      }
    \caption{Constraint $C_1$ for Jarrell's CSP on OM}
    \label{fig:jarrell}
  \end{center}
\end{figure}

We represent graphically (fig.\ref{fig:jarrell}) 
the constraint $C_1$. We use 
the \textit{x$\rightarrow$dx} and \textit{motives-occurs=}  boxes to 
fix the number of occurrences of each motive over the intervals of the output sequence.

Jarrell also proposes in \cite{jarrell} to consider absolute intervals and octaviation for the chords,
the limits and the motives. For instance, using
absolute intervals, an interval $V_{i+1} - V_i$ is equal to $V_{i} - V_{i+1}$ and using octaviation, a pitch G4 is equivalent to G1,G2,G5, etc. Finally, he also proposes to have specific motives and chords for each segment of the output sequence, according to a user-defined segmentation. For simplicity, we do not present those constraints in this paper. However, a complete model of this problem can be found at \textit{Gelisp} website.



\section{Concluding Remarks and Future Work}
We presented a library for Common Lisp and OM providing a variety of constraints and search engines. \textit{Gelisp} provides graphical boxes to represent 
some constraints and search strategies. \textit{Gelisp} abstracts minor details that are not necessary
for musicians and mathematicians.

It would be pretentious to conclude that we can easily model any musical CSP using \textit{Gelisp}  graphical interface,
or using the command-line interface. However, we can model a variety of problems
using \textit{Gelisp}  in a simple way taking advantage of the state-of-the-art propagators and search
engines provided by Gecode.

An approach related to CSPs is concurrent constraint programming, a family of process calculi often used to model musical interactions problems. 
Process calculi has been applied to the modeling of interactive music systems
 \cite{is-chapter,tdcr14,ntccrt,cc-chapter,torophd,torobsc,Toro-Bermudez10,Toro15,ArandaAOPRTV09,tdcc12,toro-report09,tdc10,tdcb10,tororeport} 
 and ecological systems \cite{PT13, TPSK14, PTA13, mean-field-techreport}.

In future works, we will explore a bigger sample
of musical CSPs and their representation using global 
constraints.
In addition, the idea of representing CSPs and their search strategies with business rules from \textit{Rules2Cp} \cite{rules4cp} can be extended to generate a musical CSP based on musical rules.


\section{ACKNOWLEDGMENTS}
Thanks to Moreno Andreatta, Jean Bresson, Serge Lemouton, Killian Sprotte, and Guido Tack for their valuable comments when developing \textit{Gelisp}.
Thanks to Carlos Toro and Jorge P\'{e}rez for their remarks on this paper.

\balance

\let\oldbibliography\thebibliography
\renewcommand{\thebibliography}[1]{%
  \oldbibliography{#1}%
  \setlength{\itemsep}{0pt}%
}

\bibliography{gelisp}

\end{document}